\documentclass{article}

\PassOptionsToPackage{numbers, compress}{natbib}
\usepackage[dblblindworkshop, final]{neurips_2025}
\workshoptitle{Generative AI in Finance}

\usepackage[utf8]{inputenc} % allow utf-8 input
\usepackage[T1]{fontenc}    % use 8-bit T1 fonts
\usepackage[colorlinks=true,linkcolor=black,citecolor=black,urlcolor=black]{hyperref}
\usepackage{url}            % simple URL typesetting
\usepackage{booktabs, multirow}       % professional-quality tables
\usepackage{amsfonts}       % blackboard math symbols
\usepackage{nicefrac}       % compact symbols for 1/2, etc.
\usepackage{microtype}      % microtypography
\usepackage{xcolor}         % colors
\usepackage[most]{tcolorbox}
\usepackage{longtable}
\usepackage{threeparttable}

\definecolor{promptbg}{RGB}{248, 249, 250}
\definecolor{promptborder}{RGB}{108, 117, 125}
\definecolor{prompttext}{RGB}{33, 37, 41}

\title{Are Foundation Models Useful for Bankruptcy Prediction?}

\author{%
  Marcin Kostrzewa$^{1}$, Oleksii Furman$^{1}$, Roman Furman$^3$, \\ \textbf{Sebastian Tomczak}$^1$, \textbf{Maciej Zięba}$^{1,2}$ \\
  $^1$Wrocław University of Science and Technology, $^2$Tooploox, $^3$Opera \\ 
  E-mail: \texttt{marcin.kostrzewa@pwr.edu.pl}
}

\begin{document}

\maketitle

\begin{abstract}
Foundation models have shown promise across various financial applications, yet their effectiveness for corporate bankruptcy prediction remains systematically unevaluated against established methods. We study bankruptcy forecasting using Llama-3.3-70B-Instruct and TabPFN, evaluated on large, highly imbalanced datasets of over one million company records from the Visegrád Group. We provide the first systematic comparison of foundation models against classical machine learning baselines for this task. Our results show that models such as XGBoost and CatBoost consistently outperform foundation models across all prediction horizons. LLM-based approaches suffer from unreliable probability estimates, undermining their use in risk-sensitive financial settings. TabPFN shows mixed results --- underperforming on ROC-AUC but competitive on $F_1$-score --- while requiring substantial computational resources that cannot be justified by its performance. These findings suggest that, despite their generality, current foundation models remain less effective than specialized methods for bankruptcy forecasting.
\end{abstract}

\section{Introduction}

Foundation models such as large language models (LLMs) are increasingly applied across various financial domains, as documented in recent survey papers \cite{Lee2024ACR, Nie2024ASO, zhao2024revolutionizingfinancellmsoverview}. Applications span sentiment analysis, algorithmic trading, fraud detection, credit rating, and other financial tasks. Their appeal lies mainly in the ability to provide solutions without extensive task-specific engineering.

One of the most extensively studied financial prediction problems is corporate bankruptcy\footnote{Throughout the paper we will use terms \textit{default}, \textit{financial distress}, and \textit{bankruptcy} interchangeably.} forecasting \cite{gupta2018empirical, hosaka2019bankruptcy, jiang2018corporate, wang2024cost, zou2022business}. Accurate default prediction is of practical relevance to investors, regulators, and policymakers, while the task's structured nature and requirement for reliable probability estimates present unique challenges for general-purpose models. While foundation models have been applied to various financial risk tasks, systematic evaluation against classical methods for bankruptcy prediction, particularly on large-scale structured datasets, remains unexplored.

In this work, we address this gap by applying two foundation models --- Llama-3.3-70B-Instruct \cite{llama3_3_modelcard} for textual inputs and TabPFN  \cite{Hollmann2025} for tabular data --- to corporate default forecasting. We evaluate them on highly imbalanced datasets of financial records from the Visegrád Group, containing hundreds of thousands of samples. While recent work has addressed TabPFN's scalability limitations \cite{Liu2025TabPFNUA, Ye2025ACL}, these solutions remain untested for financial prediction tasks against domain-optimized classical methods. We extensively compare these approaches against classical machine learning baselines and analyze the reliability of foundation model outputs for structured financial prediction.

Our contributions can be summarized as follows:
\begin{itemize}
    \item We provide the first systematic comparison of foundation models (LLMs and TabPFN) against classical ML baselines for bankruptcy prediction on large-scale structured financial data.
    \item We show that LLM self-reported probability estimates are poorly calibrated and discretized, undermining their reliability for risk assessment, and that TabPFN incurs substantial computational overhead without commensurate performance gains.
\end{itemize}

\section{Related Work}

\paragraph{Foundation models in finance.} Recent work has demonstrated foundation model capabilities across financial domains, from sentiment analysis and time series forecasting \cite{chen2025survey,Lee2024ASO, Nie2024ASO} to risk prediction. FinPT applies LLMs to personal financial risk prediction \cite{Yuwei2023}, while time series foundation models like FinCast and Kronos show strong performance for market forecasting \cite{shi2025kronosfoundationmodellanguage, zhu2025fincastfoundationmodelfinancial}. For bankruptcy prediction specifically, multimodal approaches have been explored using textual and numerical data from US markets \cite{arno2024numbers, mancisidor2022multimodal}, but systematic evaluation against classical methods on structured data remains limited.

\paragraph{Bankruptcy prediction methods.} Traditional machine learning approaches dominate bankruptcy prediction, with gradient boosting methods (XGBoost, CatBoost) and neural networks consistently achieving strong performance on financial datasets \cite{ben2023bankruptcy,jabeur2021catboost, zou2022business}. Recent studies have incorporated more advanced deep learning techniques \cite{hosaka2019bankruptcy, wu2024predicting} and ensemble methods \cite{liu2025ensemble}. Despite foundation model advances in other financial areas, no systematic comparison exists evaluating their effectiveness against these established approaches for bankruptcy prediction on large-scale datasets.

\section{Data and methodology}\label{sec:methodology}

\subsection{Data}

We construct five datasets based on over one million financial statement records from 2006-2021 of companies from the V4 group (Czech Republic, Hungary, Poland, Slovakia). The data is a part of our ongoing research and will be described in detail in a separate publication. 

Each dataset corresponds to a different prediction horizon: immediate bankruptcy risk (0 years ahead) up to four years ahead. The target variable is binary, indicating whether a company will experience financial distress as defined by a set of financial ratio thresholds. This multi-horizon approach enables systematic evaluation of model performance as prediction difficulty increases with longer time periods. Table~\ref{tab:datasets} summarizes the five datasets, showing the expected decline in both sample size and bankruptcy cases as prediction horizons extend. The severe class imbalance (bankruptcy rates below 1\%) reflects real-world conditions and necessitates careful evaluation using metrics appropriate for imbalanced classification. 

Detailed feature descriptions and target variable construction are provided in Appendix \ref{app:dataset}.

\begin{table}[ht]
\centering
\caption{Overview of datasets and prediction tasks.}
\label{tab:datasets}
\begin{tabular}{l r r r}
\toprule
\textbf{Prediction Horizon} & \textbf{Total Instances} & \textbf{Bankruptcy} & \textbf{Non-bankruptcy}  \\
\midrule
 0 years (current)   & 1,000,087 & 3,587 & 996,500 \\
 1 year ahead        &   996,500 & 3,054 & 993,446  \\
 2 years ahead       &   898,692 & 2,374 & 896,318\\
 3 years ahead       &   793,234 & 1,896 & 791,338 \\
 4 years ahead       &   700,041 & 1,485 & 698,556 \\
\bottomrule
\end{tabular}
\end{table}

\paragraph{Evaluation subsets.}
For each prediction horizon, all models are evaluated on the same stratified test subset of 20,000 samples to ensure computational feasibility and fair comparison across methods. The remaining data is split into training and validation sets, with validation data used for hyperparameter optimization and probability threshold calibration.

\subsection{Methods}

To evaluate foundation models against classical approaches for bankruptcy prediction, we employ both a LLM and a tabular foundation model alongside established baselines. For each dataset, we report ROC-AUC and $F_1$-score, with the latter being particularly important for highly imbalanced data. The decision threshold is calibrated on the validation dataset by choosing the value maximizing $F_1$-score.  Detailed results for additional metrics are provided in Appendix \ref{app:complete-results}.

\paragraph{Foundation models}

We evaluate two representative foundation models, each addressing different data modalities. For textual input processing, we use Llama-3.3-70B-Instruct \cite{llama3_3_modelcard}, serializing company features into natural-language prompts that request binary labels and self-reported probabilities. We test both zero-shot prediction and in-context learning (ICL) with 20 examples (10 positive, 10 negative). We access Llama-3.3 via API, effectively treating it as a closed model since we cannot examine weights or internal representations.

For direct tabular processing, we utilize TabPFN \cite{Hollmann2025}, a transformer-based model designed for small tabular datasets. Since TabPFN is optimized for datasets under 10{,}000 samples while our datasets contain hundreds of thousands of records, we implement a \textit{partition-then-predict} approach where a decision tree partitions the feature space and TabPFN processes smaller leaf subsets. Detailed prompt design and scaling approaches are described in Appendices \ref{app:llama} and \ref{app:tabpfn}.

\paragraph{Classical baselines.}
We compare foundation models against five established methods: logistic regression, multi-layer perceptron, XGBoost, LightGBM, and CatBoost. We follow a standard training protocol with hyperparameter optimization via grid search on validation data. Hyperparameters are selected to maximize $F_1$-score, prioritizing performance on the minority class. Complete hyperparameter grids and training specifications are provided in Appendix \ref{app:classical}.

\section{Experiments and results}\label{sec:experiments}

The results indicate that classical methods consistently outperform foundation models across all prediction horizons. Table~\ref{tab:results-roc-ap} shows that traditional ML methods maintain stable performance across all time horizons, with ROC-AUC scores declining gradually from $0.99+$ at $h=0$ to $0.85$--$0.89$ at $h=4$.

Foundation models show weaker performance. TabPFN achieves $0.987$ ROC-AUC at $h=0$, declining to $0.771$ at $h=4$, underperforming all standard ML methods across most horizons. While TabPFN outperforms MLP and LR on $F_1$-scores at most horizons, it still trails behind gradient boosting. LLM-based approaches exhibit the weakest performance in terms of ROC-AUC. The $F_1$-scores reveal even larger gaps on this imbalanced dataset, with strong classical methods achieving $0.024$--$0.069$ at $h=4$ compared to foundation models' $0.012$--$0.024$.

\begin{figure}[ht]
    \centering
    \includegraphics[width=0.6\linewidth]{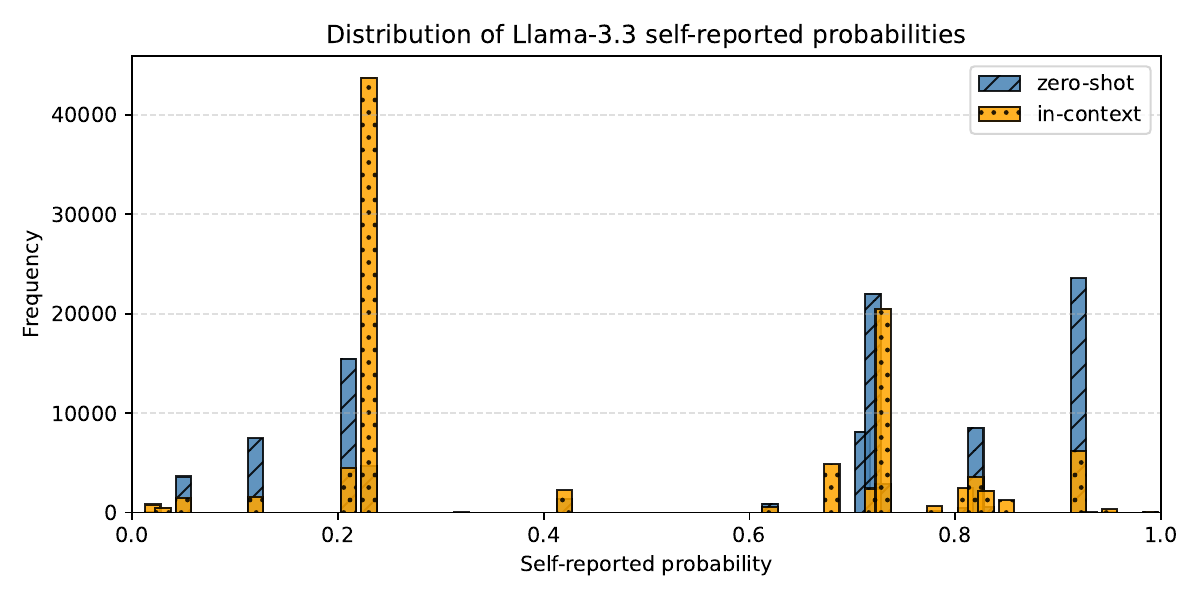}
    \caption{Distribution of self-reported classification output probabilities for Llama-3.3 in both zero-shot and in-context learning variants across all datasets.}
    \label{fig:llama-probs-dict}
\end{figure}

Beyond overall performance gaps, LLMs exhibit specific reliability issues that undermine their practical utility. The self-reported probability estimates are poorly calibrated and discretized, as shown in Figure~\ref{fig:llama-probs-dict}. The model outputs cluster around fixed values ($0.1$, $0.2$, $0.7$, $0.9$) rather than providing smooth probability distributions, making them unsuitable for risk assessment applications.

\begin{table}[ht]
\caption{Performance across prediction horizons (ROC-AUC / $F_1$-score).}
\label{tab:results-roc-ap}
\centering
\scriptsize
\begin{tabular}{lccccc}
\toprule
Model & $h=0$ & $h=1$ & $h=2$ & $h=3$ & $h=4$ \\
\midrule
XGBoost & $\mathbf{0.996}/\mathbf{0.465}$ & $\mathbf{0.968}/0.200$ & $\mathbf{0.894}/0.198$ & $0.896/0.055$ & $\mathbf{0.891}/0.024$ \\
CatBoost & $\mathbf{0.996}/0.431$ & $0.965/\mathbf{0.238}$ & $0.886/0.161$ & $\mathbf{0.901}/0.067$ & $0.883/0.062$ \\
LightGBM & $\mathbf{0.996}/0.459$ & $0.965/0.229$ & $0.878/\mathbf{0.180}$ & $0.888/0.060$ & $0.877/\mathbf{0.069}$ \\
MLP  & $0.994/0.404$ & $0.959/0.157$ & $0.848/0.059$ & $0.895/\mathbf{0.071}$ & $0.877/0.021$ \\
LR & $0.983/0.167$ & $0.952/0.148$ & $0.853/0.064$ & $0.858/0.045$ & $0.850/0.012$ \\
TabPFN & $0.987/0.400$ & $0.951/0.196$ & $0.800/0.131$ & $0.823/0.063$ & $0.771/0.024$ \\
Llama-3.3 & $0.945/0.141$ & $0.914/0.091$ & $0.796/0.020$ & $0.823/0.010$ & $0.782/0.012$ \\
Llama-3.3 (ICL) & $0.966/0.114$ & $0.932/0.064$ & $0.817/0.022$ & $0.807/0.019$ & $0.780/0.015$ \\
\bottomrule
\end{tabular}
\end{table}

Foundation models require substantial computational resources without commensurate performance gains. Our timing analysis (Table~\ref{tab:time-stats}) demonstrates significant computational overhead for foundation models, which require specialized hardware and substantially longer processing times compared to classical methods.

\begin{table}
\centering
\small
\begin{threeparttable}
\caption{Inference timing. Each entry is the mean$\pm$std over 5 runs; we also report throughput (samples/s).\label{tab:time-stats}}
\begin{tabular}{l l l r r}
\toprule
\textbf{Model} & \textbf{Hardware} & \textbf{Batch}  &
\textbf{Time (s)} & \textbf{Throughput} \\
 &  &  & $\mu\!\pm\!\sigma$ & (samples/s) \\
\midrule
XGBoost  & CPU (M4 Pro 4 threads) & 20{,}000 &  $0.007\pm0.001$ & $2,833,570.17$ \\
TabPFN   & GPU (NVIDIA A100-40GB) & 20{,}000 & $23.782\pm1.668$ & $844.77$  \\
Llama    & API (8 concurrent requests)& 20{,}000 & $5357.400\pm205.563$ & $3.739$  \\
\bottomrule
\end{tabular}
\end{threeparttable}
\end{table}

\section{Limitations}\label{sec:limitations}

Our study's limitations primarily concern the LLM approach. We use Llama-3.3-70B-Instruct, which is not among currently available top models. Advanced reasoning models (GPT-5-thinking \cite{openai2025gpt5}, Claude Opus 4.1 \cite{anthropic2025claude41}, DeepSeek-V3.1 \cite{deepseekai2024}, Qwen3 \cite{qwen3technicalreport}) may deliver stronger performance. Additionally, API-level access restricts us to returned probability estimates; direct access to weights and logits could yield more reliable probabilities.

\section{Conclusions}

In this study, we conducted the first evaluation of foundation models for corporate bankruptcy prediction. Using a large-scale dataset of Central European companies, we compared Llama-3.3 and TabPFN with established machine learning methods.

Our findings indicate that classical machine learning approaches maintain advantages over general-purpose foundation models in structured financial prediction. The performance gaps we observed, spanning accuracy, computational efficiency, and probability calibration, were not marginal but substantial, suggesting that foundation models are not an adequate choice for such task.

Most critically, we identified that LLM probability estimates exhibit a degenerate distribution, clustering around discrete values rather than providing the reliable confidence measures essential for financial decision-making. The computational overhead of foundation models without corresponding performance benefits further undermines their business case. While TabPFN demonstrates competitive $F_1$-scores on this imbalanced dataset, outperforming simpler baselines at most horizons, it underperforms on ROC-AUC and requires specialized hardware that cannot be justified when gradient boosting methods consistently deliver superior results on standard hardware.

These findings suggest that current foundation models are not yet a viable replacement for specialized machine learning approaches in bankruptcy forecasting. Nonetheless, our work highlights important challenges and points toward directions for improvement. Future research should explore LLMs with reasoning capabilities, models with accessible weights and logits, and hybrid multimodal approaches that combine textual and numerical financial data.

\bibliography{bibliography}

%%%%%%%%%%%%%%%%%%%%%%%%%%%%%%%%%%%%%%%%%%%%%%%%%%%%%%%%%%%%

\appendix

\section{Acknowledgements}
The research in this paper has been partially supported by the National Science Centre (NCN, Poland), under Grant no. 2020/39/D/HS4/02384 and  under Grant no. 2024/55/B/ST6/02100.

We would like to also gratefully acknowledge the Polish high-performance computing infrastructure PLGrid (HPC Center: ACK Cyfronet AGH) for providing computer facilities and support within computational grant no. PLG/2025/018494.

We would also like to thank CLARIN-PL for granting access to their infrastructure and API services.

\section{Dataset details}\label{app:dataset}
\subsection{Data Source and Coverage}

Our evaluation utilizes a comprehensive financial dataset sourced from the Emerging Markets Information Service (EMIS) database, covering companies across four Central European countries: Poland, Hungary, Slovakia, and the Czech Republic. The dataset spans the period from 2006 to 2021, providing a robust foundation for financial distress prediction analysis.

The dataset comprises 203{,}900 unique companies with a total of 1{,}106{,}879 company-year observations distributed as follows: Poland (628{,}499 entries), Hungary (358{,}486 entries), Slovakia (62{,}141 entries), and Czech Republic (57{,}753 entries). Each record represents a company's financial data for a specific year.

\subsection{Financial Distress Definition and Classification Task}

To construct the target variable for our predictive models, we defined financial default based on a set of financial indicators from a company's last available annual report. A company was labeled as defaulted if it simultaneously met three criteria in its final reporting year: a negative equity to total assets (\textit{equity/total\_assets} $< 0$), negative EBITDA relative to total assets (\textit{EBITDA/total\_assets} $< 0$), and a current ratio below 0.6 (\textit{current\_assets/current\_liabilities} $< 0.6$). Companies whose last available report was for the year 2021 were excluded from being labeled as distressed as we collected the data up to 2021 and we lacked the required information. For a task of predicting default 4 years in advance, we took companies that met the default criteria, removed their data for the final 4 years, and then assigned a positive label to the new final year of data for each of these firms. All other company-year observations in the datasets were assigned a negative label. This process resulted in a dataset tailored for predicting financial default at a horizon of 4 years before the observed default. In a similar manner, we construct datasets with prediction horizons $h=1, 2, 3$.

\subsection{Feature Set and Financial Indicators}
The dataset contains 131 features encompassing various categories of financial indicators. A detailed features list with descriptions is presented in Table~\ref{tab:all_features}.

\begin{longtable}{|p{0.13\textwidth}|p{0.75\textwidth}|}
\caption{The complete set of features considered in the classification process.}
\label{tab:all_features}
\\

\hline
\textbf{Feature ID} & \textbf{Description} \\
\hline
\endfirsthead

\hline
\textbf{Feature ID} & \textbf{Description} \\
\hline
\endhead

\hline
\endfoot

\hline
\endlastfoot

\multicolumn{2}{|l|}{\textbf{Company Identifiers \& Metadata}} \\
\hline
X1 & Country \\
\hline
X2 & Has multiple industries flag (binary) \\
\hline
X3 & Incorporation date 1 \\
\hline
X4 & Incorporation date 2 \\
\hline
X5 & Legal form (LLC, Corp, etc.) \\
\hline
X6 & NAICS 2-digit classification \\
\hline
X7 & NAICS 3-digit classification \\
\hline
X8 & Number of employees \\
\hline
X9 & Operational status (Active, Inactive, etc.) \\
\hline
X10 & Primary NAICS (encoded) \\
\hline
X11 & Secondary NAICS (encoded) \\
\hline
X12 & Sector 1 \\
\hline
X13 & State/region \\
\hline
X14 & Report year \\
\hline

\multicolumn{2}{|l|}{\textbf{Liquidity and Profitability Ratios}} \\
\hline
X15 & Cash / sales \\
\hline
X16 & Cash / total assets \\
\hline
X17 & Cash / total operating revenue \\
\hline
X18 & (Current assets - inventories - receivables) / short term liabilities \\
\hline
X19 & (Current assets - inventories) / short term liabilities \\
\hline
X20 & Current assets / sales \\
\hline
X21 & Current assets / short term liabilities \\
\hline
X22 & EBIT / equity \\
\hline
X23 & EBIT / financial costs \\
\hline
X24 & EBIT / total assets \\
\hline
X25 & EBIT / total costs \\
\hline
X26 & EBIT / total liabilities \\
\hline
X27 & EBIT / total operating revenue \\
\hline
X28 & EBITDA / fixed assets \\
\hline
X29 & EBITDA / total assets \\
\hline
X30 & EBITDA / total operating revenue \\
\hline
X31 & (Gross profit + depreciation) / total liabilities \\
\hline
X32 & Gross profit / short term liabilities \\
\hline
X33 & Gross profit / total assets \\
\hline
X34 & Gross profit / total operating revenue \\
\hline
X35 & Interest expense / revenue \\
\hline
X36 & Inventories / working capital \\
\hline
X37 & (Net profit + depreciation) / current liabilities \\
\hline
X38 & (Net profit + depreciation) / total liabilities \\
\hline
X39 & Net profit / equity \\
\hline
X40 & Net profit / fixed assets \\
\hline
X41 & Net profit / inventories \\
\hline
X42 & Net profit / total assets \\
\hline
X43 & Net profit / total operating revenue \\
\hline
X44 & Net profit / current assets \\
\hline
X45 & Operational expenses / short term liabilities \\
\hline
X46 & Operational expenses / total liabilities \\
\hline
X47 & Quick assets / sales \\
\hline
X48 & Retained profit / short term liabilities \\
\hline
X49 & Retained profit / total assets \\
\hline
X50 & Working capital (absolute value) \\
\hline
X51 & Working capital / equity \\
\hline
X52 & Working capital / fixed assets \\
\hline
X53 & Working capital / sales \\
\hline
X54 & Working capital / total assets \\
\hline
X55 & Working capital / total liabilities \\
\hline
X56 & Working capital / total operating revenue \\
\hline
X57 & Cash flow / sales \\
\hline
X58 & Cash flow / total debt \\
\hline
X59 & Loss flag (Net Profit (Loss) for the Period $< 0$) \\
\hline

\multicolumn{2}{|l|}{\textbf{Turnover and Cycle Ratios}} \\
\hline
X60 & Cash conversion cycle (days) \\
\hline
X61 & Inventories / total operating revenue \\
\hline
X62 & Operating cycle (days) \\
\hline
X63 & Operating expenses / sales \\
\hline
X64 & Receivables turnover days \\
\hline
X65 & Revenue / current assets \\
\hline
X66 & Revenue / long term liabilities \\
\hline
X67 & Revenue / total liabilities \\
\hline
X68 & Short term liabilities turnover days \\
\hline
X69 & Total operating revenue / fixed assets \\
\hline
X70 & Total operating revenue / inventories \\
\hline
X71 & Total operating revenue / receivables \\
\hline
X72 & Total operating revenue / short term liabilities \\
\hline
X73 & Total operating revenue / total assets \\
\hline

\multicolumn{2}{|l|}{\textbf{Solvency and Capital Structure Ratios}} \\
\hline
X74 & Constant capital / fixed assets \\
\hline
X75 & Constant capital / total assets \\
\hline
X76 & Current assets / total liabilities \\
\hline
X77 & Current assets / total operating revenue \\
\hline
X78 & Current liabilities / total liabilities \\
\hline
X79 & Current liabilities / current assets \\
\hline
X80 & Current liabilities / equity \\
\hline
X81 & Short term liabilities / total assets \\
\hline
X82 & (Equity - share capital) / fixed assets \\
\hline
X83 & Equity / fixed assets \\
\hline
X84 & Equity / long term liabilities \\
\hline
X85 & Equity / sales \\
\hline
X86 & Equity / total assets \\
\hline
X87 & Equity / total liabilities \\
\hline
X88 & Equity ratio classification \\
\hline
X89 & Fixed assets / long term liabilities \\
\hline
X90 & Fixed assets / total assets \\
\hline
X91 & (Inventories + receivables) / equity \\
\hline
X92 & Inventory / current liabilities \\
\hline
X93 & Long term liabilities / current assets \\
\hline
X94 & Long term liabilities / equity \\
\hline
X95 & (Total liabilities - cash) / EBITDA \\
\hline
X96 & (Total liabilities - cash) / total operating revenue \\
\hline
X97 & Total liabilities / total assets \\
\hline
X98 & Insolvency flag (Total liabilities $>$ Total assets) \\
\hline

\multicolumn{2}{|l|}{\textbf{Growth Ratios (Year-over-Year)}} \\
\hline
X99 & Current assets growth (YoY) \\
\hline
X100 & Inventories growth (YoY) \\
\hline
X101 & Net profit growth (YoY) \\
\hline
X102 & Operating profit growth (YoY) \\
\hline
X103 & Operating revenue growth (YoY) \\
\hline
X104 & Receivables growth (YoY) \\
\hline
X105 & Short term liabilities growth (YoY) \\
\hline
X106 & Total assets growth (YoY) \\
\hline

\multicolumn{2}{|l|}{\textbf{Additional and Derived Indicators (Size and Macro)}} \\
\hline
X107 & Logarithm of current assets \\
\hline
X108 & Logarithm of (net profit / GDP) \\
\hline
X109 & Logarithm of (operating profit / GDP) \\
\hline
X110 & Logarithm of (revenue / GDP) \\
\hline
X111 & Logarithm of total liabilities \\
\hline
X112 & Logarithm of total assets \\
\hline
X113 & Logarithm of (total assets / GDP) \\
\hline
X114 & Logarithm of total operating revenue \\
\hline

\multicolumn{2}{|l|}{\textbf{Sector-Relative Indicators}} \\
\multicolumn{2}{|p{0.75\textwidth}|}{\textit{Represents the difference between the company's ratio and the industry sector average for that year.}} \\
\hline
X115 & Cash conversion cycle (sector-relative) \\
\hline
X116 & (Current liabilities $\times 365$) / revenue (sector-relative) \\
\hline
X117 & Current assets / current liabilities (sector-relative) \\
\hline
X118 & EBITDA margin (sector-relative) \\
\hline
X119 & (Inventories $\times 365$) / revenue (sector-relative) \\
\hline
X120 & Net profit / absolute equity (sector-relative) \\
\hline
X121 & Net profit / assets (sector-relative) \\
\hline
X122 & Net profit / current assets (sector-relative) \\
\hline
X123 & Net profit / fixed assets (sector-relative) \\
\hline
X124 & Net profit / sales (sector-relative) \\
\hline
X125 & Operating cycle (sector-relative) \\
\hline
X126 & (Receivables $\times 365$) / revenue (sector-relative) \\
\hline
X127 & Revenue / assets (sector-relative) \\
\hline
X128 & Revenue / fixed assets (sector-relative) \\
\hline
X129 & Short-term financial assets / current liabilities (sector-relative) \\
\hline
X130 & Short-term receivables investments / current liabilities (sector-relative) \\
\hline
X131 & Working capital / assets (sector-relative) \\
\bottomrule
\end{longtable}

\section{Model training and evaluation details}

\paragraph{Data preprocessing.}

For each dataset, we impute missing numeric values with training dataset medians and standardize using training dataset statistics (scaling is not performed for data fetched by LLM). Categorical columns such as \emph{Country}, \emph{State/Region}, and \emph{Legal form} are label-encoded.

\subsection{Llama-3.3}\label{app:llama}

\paragraph{Prompt design.}

In our prompt template, we provide guidance for bankruptcy prediction and impose a financial analyst persona on LLM. The template includes detailed explanations of financial ratio categories, risk interpretation guidelines, and strict output formatting requirements. When using in-context learning, an \texttt{**Examples**} section with 20 examples is inserted before the target company details.

The complete prompt structure is shown below.

\begin{tcolorbox}[
    enhanced,
    colback=promptbg,
    colframe=promptborder,
    arc=4pt,
    outer arc=1pt,
    boxrule=1pt,
    left=8pt,
    right=8pt,
    top=8pt,
    bottom=8pt,
    fontupper=\small\ttfamily,
    colupper=prompttext,
    breakable,
    title={\textbf{LLM Prompt Template for Bankruptcy Prediction}},
    fonttitle=\normalfont\bfseries,
    colbacktitle=white,
    coltitle=black,
    attach boxed title to top center={yshift=-2mm},
    boxed title style={boxrule=0pt,arc=2pt}
]

You are a financial analyst specializing in bankruptcy prediction for companies. Analyze the provided financial data and predict [TIMEFRAME] bankruptcy risk.

\vspace{2mm}

**Task**: Based on the financial metrics below, predict if this company will go bankrupt within the next [X] years.

\vspace{2mm}

**Prediction Timeframe**: This is a [X]-year ahead prediction. [TIMEFRAME-SPECIFIC GUIDANCE]

\vspace{2mm}

**Key Financial Ratios Explained**:

  - **Liquidity Ratios**: Measure ability to pay short-term debts
  
  - Current\_assets/ current\_liabilities: Current ratio (>1.0 = good liquidity)
  
  - Current\_assets-inventories/ current\_liabilities: Acid-test ratio (removes 
  less liquid inventories)
  
  - Current\_assets-inventories-receivables/ current\_liabilities: Quick ratio (most conservative liquidity measure)
  
  - Working\_capital: Current assets minus current liabilities (positive = good)
  
  - Working\_capital/total\_assets: Working capital efficiency (higher = better liquidity management)
  
  - Cash/total\_assets: Cash position relative to total assets (higher = safer)
  
\vspace{1mm}
- **Profitability Ratios**: Measure earnings performance

  - Net\_profit/total\_assets: Return on assets (ROA, higher = better)
  
  - EBIT/total\_assets: Operating return on assets (excludes financial structure effects)
  
  - EBITDA/total\_assets: Cash-based return on assets (excludes depreciation)
  
  - Net\_profit/total\_operating\_revenue: Net profit margin (higher = better)
  
  - EBIT/total\_operating\_revenue: Operating margin (higher = better)
  
  - EBITDA/total\_operating\_revenue: EBITDA margin (cash flow efficiency)
  
  - EBIT/equity: Return on equity (higher = better)
  
  - EBIT/total\_costs: Operating efficiency ratio
  
  - Net\_profit/fixed\_assets: Asset utilization efficiency
  
  - EBITDA/fixed\_assets: Cash generation from fixed assets
  
\vspace{1mm}
- **Leverage Ratios**: Measure debt levels and financial risk

  - EBIT/total\_liabilities: Earnings coverage of total debt (higher = better)
  
  - Net\_profit+depreciation/total\_liabilities: Cash flow coverage of debt
  
  - Total\_liabilities/total\_assets: Debt-to-asset ratio (lower = better)
  
  - Equity/total\_assets: Equity ratio (higher = better)
  
  - Equity/fixed\_assets: Equity financing of fixed assets (higher = better)
  
  - Long\_term\_liabilities/equity: Long-term debt burden
  
  - Current\_liabilities/current\_assets: Short-term debt pressure (lower = better)
  
\vspace{1mm}
- **Efficiency Ratios**: Measure operational performance

  - Operating\_cycle: Days to convert inventory to cash (lower = better)
  
  - Cash\_conversion\_cycle: Net days to convert investments to cash
  
  - Receivables\_turnover\_days: Days to collect receivables (lower = better)
  
\vspace{1mm}
- **Growth Metrics**: Measure company expansion

  - Operating\_revenue\_growth: Revenue growth rate
  
  - Total\_assets\_growth: Asset growth rate
  
  - Net\_profit\_growth: Profit growth rate
  
\vspace{1mm}
- **Asset Composition**: Measure asset structure and efficiency

  - Fixed\_assets/total\_assets: Asset structure (higher = more capital intensive)
  
  - Working\_capital/fixed\_assets: Working capital relative to fixed investment
  
  - Current\_assets/total\_liabilities: Asset coverage of liabilities
  
\vspace{1mm}
- **Risk Flags**: Binary indicators

  - Insolvency\_flag: 1 if company is technically insolvent
  
  - Loss\_flag: 1 if company has consecutive losses
  
\vspace{2mm}
**Bankruptcy Risk Indicators**:

- **High Risk**: Negative working capital, low liquidity ratios (<1.0), high debt ratios (>0.7), declining profits, insolvency/loss flags

- **Medium Risk**: Declining growth, moderate debt levels (0.4-0.7), industry volatility, operational issues

- **Low Risk**: Strong liquidity (>1.5), positive growth, low debt (<0.4), consistent profitability, stable industry

\vspace{2mm}
**Ratio Interpretation Guidelines**:

- Current ratio < 1.0: Severe liquidity problems

- Quick ratio < 1.0: Potential liquidity problems

- Working capital/total\_assets < 0: Negative working capital (high risk)

- Debt-to-asset ratio > 0.7: High financial risk

- EBIT/total\_liabilities < 0.1: Poor debt coverage

- Fixed\_assets/total\_assets > 0.8: High capital intensity (industry dependent)

- Negative growth rates: Declining business performance

- Operating cycle > 120 days: Inefficient operations

- Loss flags = 1: Immediate bankruptcy risk

\vspace{2mm}
**Company Structure Features**:

- **Industry Codes**: NAICS classification system

  - primary\_naics\_encoded: Main industry code (higher numbers = more specific industries)
  
  - naics\_2digit: Broad sector (11-99, e.g., 23=Construction, 31-33=Manufacturing)
  
  - naics\_3digit: Subsector (e.g., 236=Construction of Buildings)
  
  - has\_multiple\_industries: 1 if company operates in multiple industries (higher risk)
  
  - secondary\_naics\_encoded: Secondary industry if diversified
  
- **Incorporation Date**: 

  - incorporation\_date\_1: 0-2y, 3-4y, 5-24y, >24y
  
  - incorporation\_date\_2: 0-1y, 1-2y, 3-5y, 6-9y, 10-19y, >19y
  
- **Operational Status**: Liquidation, Under Legal Investigation, Closed, Active

\vspace{2mm}
**Response Format**: Respond with exactly two numbers separated by a comma:
[prediction],[probability]

\vspace{1mm}
Where:
- prediction: 1 if the company will likely go bankrupt, 0 if not
- probability: a decimal between 0.0 and 1.0 representing the probability of bankruptcy

\vspace{1mm}
Example responses:

- "1,0.85" (high bankruptcy risk)

- "0,0.15" (low bankruptcy risk)

- "1,0.65" (moderate-high bankruptcy risk)

\vspace{1mm}
RETURN ONLY THE REQUIRED NUMBERS, NO OTHER TEXT

\vspace{3mm}

**Examples**:

\vspace{1mm}

Example 1:
Company Info: country=Hungary, state=Bacs-Kiskun, number\_of\_employees=1-9 employees, legal\_form=Limited Liability Partnership, primary\_naics\_encoded=42512, naics\_2digit=42, naics\_3digit=425, has\_multiple\_industries=Single Industry, secondary\_naics\_encoded=0, sector\_1=Wholesale Trade, year=2,019.00

Liquidity: Working\_capital=-25.170, Cash/total\_assets=0.000, Inventories/working\_capital=0.000...

[Additional financial metrics...]

Risk Flags: Insolvency\_flag=1.000, Loss\_flag=0.000
Prediction: 0

\vspace{2mm}

Example 2:
Company Info: country=Poland, state=Podlaskie, number\_of\_employees=10-49 employees, legal\_form=Limited Liability Company...

[Financial metrics...]

Prediction: 0

\vspace{1mm}

...

\vspace{3mm}

**Now analyze this company**:

Company Info: country=Poland, state=Lodzkie, number\_of\_employees=10-49 employees, legal\_form=Limited Liability Partnership, primary\_naics\_encoded=236.0, naics\_2digit=23.0, naics\_3digit=236.0...

[Complete set of financial indicators...]

Prediction:

\end{tcolorbox}

\paragraph{In-context learning examples selection.}

Our procedure for selecting $2k$ examples for in-context learning looks as follows:
\begin{itemize}
    \item we train a proxy XGBoost model on the training data and score the validation dataset to obtain probabilities,
    \item we choose $k$ samples from bankruptcy class with lowest probabilities and $k$ samples from non-bankruptcy class with highest probabilities.
\end{itemize}

In other words, we try to choose \textit{hard} samples in terms of prediction for in-context learning. For all tasks, we choose $k=10$.

\paragraph{Calibrating threshold}
Threshold for calculating metrics is selected by maximizing $F_1$ score on a dataset of 5{,}000 samples sub-sampled with stratification from the validation dataset.

\subsection{TabPFN approaches}\label{app:tabpfn}
We implement two strategies to handle our large-scale datasets with TabPFN, which is optimized for datasets under 10,000 samples.

\paragraph{Decision Tree Partitioning (TabPFN-DT)}

We partition the entire training dataset using shallow decision tree, setting the minimum number of samples required to split an internal node to 10,000. The decision tree partitions the training set into smaller, more manageable subsets. During inference, a test instance is first passed through the decision tree to a leaf node and then predicted by the corresponding TabPFN model.

\paragraph{Bootstrap Ensemble (TabPFN-Ensemble)}

We also implement a bootstrap ensemble approach where we iteratively sample
$m$ subsets, each containing $n < N$ randomly selected samples. For each subset, we leverage TabPFN's ability to handle the reduced dataset size to obtain predictions. This divide-and-conquer strategy aggregates outputs using majority voting for classification. The approach's performance is determined by the number of datasets bootstrapped.

\paragraph{Performance comparison}
Table~\ref{tab:tabpfn-comparison} compares the performance of both TabPFN approaches for the 4-year horizon prediction. Based on the obtained results, we use the TabPFN-DT approach in the main text analysis.

\begin{table}[ht]
\caption{TabPFN approaches performance comparison for $h=4$ (ROC-AUC / $F_1$-score).}
\label{tab:tabpfn-comparison}
\centering
\scriptsize
\begin{tabular}{lcc}
\toprule
Model & ROC-AUC & $F_1$-score\\
\midrule
TabPFN-DT           & 0.771 & $\mathbf{0.024}$ \\
TabPFN-Ensemble-8   & 0.797 & 0.018
 \\
TabPFN-Ensemble-16  & 0.794 & 0.013
 \\
TabPFN-Ensemble-24  & $\mathbf{0.814}$ & 0.013 \\
\bottomrule
\end{tabular}
\end{table}

\paragraph{Threshold calibration}
The threshold for calculating metrics on the test dataset is selected to maximize the $F_1$-score on the validation dataset.

\subsection{Classical methods hypertuning}\label{app:classical}

\paragraph{Evaluation protocol.}
For each prediction horizon, metrics are computed on a common stratified test \emph{subset} of 20{,}000 samples (computational parity across models).
We do not use class weights or resampling; class imbalance is handled via $F_1$-score-based model selection and threshold calibration.

\paragraph{Models and search spaces.} 
Hyperparameters are selected using grid search by maximizing the $F_1$-score on the validation dataset. After selection, a decision threshold is calibrated on the validation dataset to maximize the $F_1$-score; this threshold is stored and used at test time when evaluating metrics.

The complete configuration of the search grid is given in Table \ref{tab:app-hparams}.

\begin{table}[h]
\centering
\caption{Hyperparameter ranges used in grid search.}
\label{tab:app-hparams}
\begin{tabular}{@{}ll@{}}
\toprule
\textbf{Model} & \textbf{Search space} \\
\midrule
LR & \(C \in \text{logspace}(0.001, 1, 5)\); penalty = L2 \\
MLP & hidden\_layer\_sizes \(\in \{(32,32),(64,64),(128,128),(256,256),(512,512)\}\); \\
    & \(\alpha \in \{10^{-4},10^{-3},10^{-2}\}\);\; learning\_rate\_init \(\in \{10^{-3},10^{-2}\}\) \\
XGBoost & n\_estimators 
\(\in \{100,200,500\}\);\; max\_depth \(\in \{3,5,7\}\);\; \\
        & learning\_rate \(\in \{0.01,0.05,0.1,0.2\}\);\; eval\_metric = logloss \\
LightGBM & n\_estimators \(\in \{100,200\}\);\; max\_depth \(\in \{-1,5,10\}\);\; \\
         & learning\_rate \(\in \{0.01,0.05,0.1,0.2\}\) \\
CatBoost & iterations \(\in \{100,200\}\);\; depth \(\in \{4,6,8\}\);\; learning\_rate \(\in \{0.01,0.05,0.1\}\);\; \\ & silent=True \\
\bottomrule
\end{tabular}
\end{table}

\paragraph{Software.}
Implementations rely on \texttt{scikit-learn}, \texttt{xgboost}, \texttt{lightgbm}, and \texttt{catboost}. 
All runs use \(\texttt{random\_state}=42\).

\section{Complete results}\label{app:complete-results}

Table~\ref{tab:complete_results} gathers complete results for all metrics (accuracy, precision, recall, $F_1$-score, ROC-AUC) evaluated for all models for each specific dataset.

\begin{table}[ht]
\caption{Complete performance results across all prediction horizons and metrics.}
\label{tab:complete_results}
\centering
\tiny
\begin{tabular}{llccccc}
\toprule
Prediction Horizon & Model & Accuracy & Precision & Recall & $F_1$-score & ROC-AUC \\
\midrule
\multirow{8}{*}{$h=0$}
& XGBoost & $0.995$ & $0.368$ & $0.630$ & $\mathbf{0.465}$ & $\mathbf{0.996}$ \\
& CatBoost & $0.994$ & $0.336$ & $0.603$ & $0.431$ & $\mathbf{0.996}$ \\
& LightGBM & $\mathbf{0.995}$ & $\mathbf{0.382}$ & $0.575$ & $0.459$ & $\mathbf{0.996}$ \\
& MLP & $0.993$ & $0.300$ & $0.616$ & $0.404$ & $0.994$ \\
& LR & $0.990$ & $0.123$ & $0.260$ & $0.167$ & $0.983$ \\
& TabPFN & $\mathbf{0.995}$ & $0.336$ & $0.493$ & $0.400$ & $0.987$ \\
& Llama-3.3 & $0.973$ & $0.080$ & $0.616$ & $0.141$ & $0.945$ \\
& Llama-3.3 (ICL) & $0.964$ & $0.063$ & $\mathbf{0.644}$ & $0.114$ & $0.966$ \\
\midrule
\multirow{8}{*}{$h=1$}
& XGBoost & $0.993$ & $0.150$ & $0.298$ & $0.200$ & $\mathbf{0.968}$ \\
& CatBoost & $0.993$ & $0.172$ & $\mathbf{0.386}$ & $\mathbf{0.238}$ & $0.964$ \\
& LightGBM & $\mathbf{0.994}$ & $\mathbf{0.174}$ & $0.333$ & $0.229$ & $0.965$ \\
& MLP & $0.992$ & $0.116$ & $0.246$ & $0.157$ & $0.959$ \\
& LR & $0.993$ & $0.114$ & $0.210$ & $0.148$ & $0.952$ \\
& TabPFN & $\mathbf{0.994}$ & $0.156$ & $0.263$ & $0.196$ & $0.951$ \\
& Llama-3.3 & $0.991$ & $0.064$ & $0.158$ & $0.091$ & $0.914$ \\
& Llama-3.3 (ICL) & $\mathbf{0.994}$ & $0.059$ & $0.070$ & $0.064$ & $0.932$ \\
\midrule
\multirow{8}{*}{$h=2$}
& XGBoost & $0.993$ & $\mathbf{0.145}$ & $0.309$ & $\mathbf{0.198}$ & $\mathbf{0.894}$ \\
& CatBoost & $\mathbf{0.995}$ & $\mathbf{0.145}$ & $0.182$ & $0.161$ & $0.886$ \\
& LightGBM & $0.993$ & $0.134$ & $0.273$ & $0.180$ & $0.878$ \\
& MLP & $0.992$ & $0.044$ & $0.091$ & $0.059$ & $0.848$ \\
& LR & $0.993$ & $0.049$ & $0.091$ & $0.064$ & $0.853$ \\
& TabPFN & $0.992$ & $0.094$ & $0.218$ & $0.131$ & $0.800$ \\
& Llama-3.3 & $0.804$ & $0.010$ & $\mathbf{0.727}$ & $0.020$ & $0.796$ \\
& Llama-3.3 (ICL) & $0.991$ & $0.016$ & $0.036$ & $0.022$ & $0.817$ \\
\midrule
\multirow{8}{*}{$h=3$}
& XGBoost & $0.995$ & $0.046$ & $0.067$ & $0.054$ & $0.896$ \\
& CatBoost & $0.990$ & $0.043$ & $0.156$ & $0.067$ & $\mathbf{0.901}$ \\
& LightGBM & $0.988$ & $0.036$ & $0.178$ & $0.060$ & $0.888$ \\
& MLP & $\mathbf{0.996}$ & $\mathbf{0.075}$ & $0.067$ & $\mathbf{0.071}$ & $0.895$ \\
& LR & $0.992$ & $0.030$ & $0.089$ & $0.046$ & $0.858$ \\
& TabPFN & $0.993$ & $0.044$ & $0.111$ & $0.063$ & $0.823$ \\
& Llama-3.3 & $0.602$ & $0.005$ & $\mathbf{0.889}$ & $0.010$ & $0.823$ \\
& Llama-3.3 (ICL) & $0.938$ & $0.010$ & $0.267$ & $0.019$ & $0.807$ \\
\midrule
\multirow{8}{*}{$h=4$}
& XGBoost & $\mathbf{0.996}$ & $0.025$ & $0.022$ & $0.024$ & $\mathbf{0.891}$ \\
& CatBoost & $0.988$ & $0.038$ & $0.178$ & $0.062$ & $0.883$ \\
& LightGBM & $0.990$ & $\mathbf{0.044}$ & $0.156$ & $\mathbf{0.069}$ & $0.877$ \\
& MLP & $0.991$ & $0.014$ & $0.044$ & $0.021$ & $0.877$ \\
& LR & $0.992$ & $0.008$ & $0.022$ & $0.012$ & $0.850$ \\
& TabPFN & $0.992$ & $0.016$ & $0.044$ & $0.024$ & $0.771$ \\
& Llama-3.3 & $0.783$ & $0.006$ & $\mathbf{0.600}$ & $0.012$ & $0.782$ \\
& Llama-3.3 (ICL) & $0.948$ & $0.008$ & $0.178$ & $0.015$ & $0.780$ \\
\bottomrule
\end{tabular}
\end{table}

\end{document}